%% file: zeller2024icra.tex
\documentclass[letterpaper, 10 pt, conference]{ieeeconf}
\IEEEoverridecommandlockouts    %
\input{stachnisslab-latex}

\input{stachnisslab-math}

\usepackage{siunitx}
\usepackage{booktabs}
\usepackage{graphicx}

\title{\LARGE \bf Radar Tracker: Moving Instance Tracking \\ in Sparse and Noisy Radar Point Clouds}

\author{Matthias Zeller \and Daniel Casado Herraez \and Jens Behley \and Michael Heidingsfeld \and Cyrill Stachniss%
  \thanks{Matthias Zeller and Daniel Casado Herraez are with CARIAD SE and with the Center for Robotics, University of Bonn, Germany. Jens Behley is with the Center for Robotics, University of Bonn, Germany. Michael Heidingsfeld is with CARIAD SE, Germany. Cyrill Stachniss is with the Center for Robotics, University of Bonn, Germany, and with the Lamarr Institute for Machine Learning and Artificial Intelligence, Germany.}%
}

\begin{document}
\maketitle
\thispagestyle{empty}
\pagestyle{empty}

\begin{abstract}

Robots and autonomous vehicles should be aware of what happens in their surroundings. The segmentation and tracking of moving objects are essential for reliable path planning, including collision avoidance. We investigate this estimation task for vehicles using radar sensing.
We address moving instance tracking in sparse radar point clouds to enhance scene interpretation.
We propose a learning-based radar tracker incorporating temporal offset predictions to enable direct center-based association and enhance segmentation performance by including additional motion cues. We implement attention-based tracking for sparse radar scans to include appearance features and enhance performance. The final association combines geometric and appearance features to overcome the limitations of center-based tracking to associate instances reliably.
Our approach shows an improved performance on the moving instance tracking benchmark of the RadarScenes dataset compared to the current \mbox{state of the art}.

\end{abstract}

\section{Introduction}
\label{sec:intro}

Motion planning and reliable collision avoidance of autonomous vehicles in real-world environments depend on the precise tracking of moving instances. The information on how many agents are present and the prediction of movement is crucial for path planning and pose estimation. Cameras, LiDARs, and radars provide valuable information about the surroundings, and a versatile setup of autonomous vehicles often aims to reduce critical malfunctions. Radar sensor measurements are often noisy due to multi-path propagation, sensor noise, and ego motion. However, radar sensors work under adverse weather, overcoming the limitations of cameras and LiDARs, and thus are essential for reliable perception systems. 
Additionally, radar sensors directly measure the Doppler velocity of the so-called radar detections and determine the radar cross section, which depends on the material, the geometry, and the surface of the object, which could help to identify and track moving agents precisely.
\begin{figure}[t]
  \centering
  \fontsize{9pt}{9pt}\selectfont
     \def\svgwidth{\linewidth}
     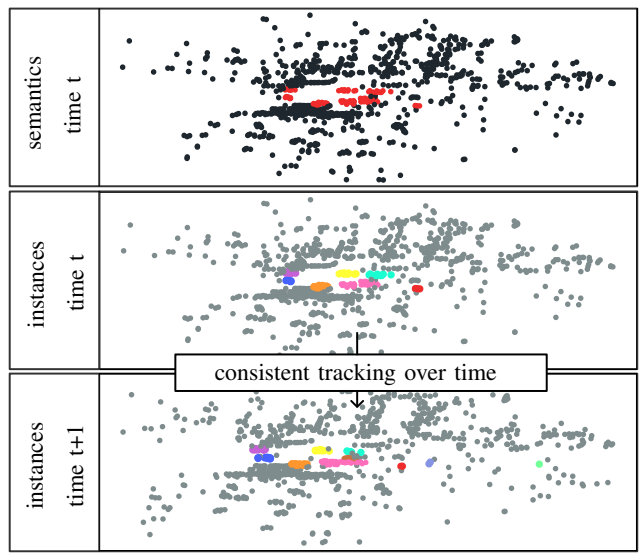
  \caption{Our method combines moving object segmentation (top), instance segmentation (middle), and tracking (bottom) to solve the 4D panoptic task of moving instance tracking from sparse radar point clouds. The corresponding colors in the middle and bottom images represent the respective tracked instances (static is grey).}
  \label{fig:motivation}
  \vspace{-0.3cm}
\end{figure}

In this paper, we elaborate on moving instance segmentation and tracking in noisy and sparse radar point clouds, as illustrated in~\figref{fig:motivation}. This requires differentiating between moving and static parts of the surroundings and consistently distinguishing instances of individual agents in the environment over time. The 4D moving object segmentation task falls within 4D panoptic segmentation~\cite{ayguen2021cvpr}. However, all moving objects belong to the moving object class without further differentiation into a more detailed separation.

Current state-of-the-art methods~\cite{chen2023cvpr,marcuzzi2022icra} often address moving instance tracking within aggregated scans and associate instances and existing tracks based on the intersection over union~(IoU) score. However, the aggregation of scans induces latency and is disadvantageous for tasks requiring immediate feedback, such as collision avoidance. Additionally, instances within sparse radar point clouds often comprise single points for which an IoU-based association is inappropriate. Other methods rely on dedicated trackers based on Kalman filters~\cite{kalman1960tasme}, which often neglect valuable appearance features~\cite{chiu2020arxiv,weng2020iros}. To extract appearance features, other approaches~\cite{chen2023cvpr,marcuzzi2022icra} voxelize the point clouds, which is particularly harmful to sparse radar data processing~\cite{zeller2023ral}.

The main contribution of this paper is a novel point-based approach that enables moving instance tracking by incorporating geometric and appearance features to accurately associate moving instances in sparse and noisy radar point clouds over time. Our approach, called Radar Tracker, utilizes neural networks and extends the prediction of a moving instance segmentation approach to derive temporal consistent tracking IDs. 
We efficiently incorporate temporal information for each point by a temporal offset prediction to enhance the segmentation and enable direct center-based tracking of moving instances. We propose an attention-based instance feature extraction network to reduce information loss and keep the appearance features of the individual instances. Furthermore, we derive attention-based association scores to extend tracking by attention. The final geometric and appearance features are combined within our data association to improve the performance of the overall estimation task.

In sum, we make three claims: First, our approach shows state-of-the-art performance for moving instance tracking in sparse and noisy radar point clouds. Second, our temporal offset prediction incorporates valuable information for segmentation and enhances tracking performance. Third, our attention-based track association overcomes the shortcomings of center-based tracking by incorporating appearance feature information. 

\section{Related Work}
\label{sec:related}

Moving instance tracking tasks can be solved as 4D panoptic segmentation~\cite{ayguen2021cvpr,kreuzberg2022eccv,zhu2023arxiv} combining moving instance segmentation and tracking. Additionally, the task benefits from multi-object tracking~\cite{weng2020iros,zhang2023cvprws}, single object tracking~\cite{hui2022eccv,xu2023cvpr}, and panoptic segmentation~\cite{hong2021cvpr,zeller2023tro}. 

Panoptic Segmentation unifies instance and semantic segmentation. There exists extensive literature, including projection-based~\cite{chen2021ral,kim2022ral,sun2022iros,qiu2022tmlr,zhang2020cvpr,huang2022eccv}, voxel-based~\cite{choy2019cvpr,mersch2022ral,zhu2021cvpr,vu2022cvpr,li2022cvpr,su2023arxiv,wang2023arxiv}, point-based~\cite{dubey2022mlwa,qi2017cvpr,qi2017nips2,schumann2018icif}, and transformer-based~\cite{marcuzzi2023ral,park2022cvpr,schult2023icra,shi2022icra,xin2022cvpr,zhang2022ijis,zhao2021iccv,zeller2023ral} approaches to solving individual tasks. However, projection-based and voxel-based approaches inherently introduce discretization artifacts and information loss, which is harmful to the targeted processing of sparse radar point clouds.

The point-based approaches directly process point clouds, keeping the spatial information intact to overcome the lossy encoding. Schumann~\etalcite{schumann2018icif} adopted this approach by aggregating multiple radar point clouds as input for PointNet++~\cite{qi2017nips2} and improved the method by adding a temporal module and additional features~\cite{schumann2020tiv}. The aggregation of scans induces latency which is disadvantageous for safety-critical tasks requiring immediate feedback. 

Recently, transformer-based networks have overcome this limitation by exploiting the self-attention mechanism~\cite{vaswani2017nips} in point cloud understanding~\cite{guo2021pct,park2022cvpr,schult2023icra,shi2022icra,xin2022cvpr,zhang2022ijis,zhao2021iccv}, which is inherently suitable to capture strong local and global dependencies and thus enable single-scan radar processing~\cite{zeller2023ral,zeller2023icra}. In our prior work~\cite{zeller2023tro}, we efficiently include the temporal information within a single scan and proposed an attention-based class-agnostic instance assignment to reliably segment moving instances. However, the trajectory and tracking information about moving agents is missing, which is required for safe autonomous mobility.

4D Panoptic Segmentation unifies instance segmentation and tracking~\cite{kim2020cvpr} and thus incorporates spatial and temporal information about the environment. Only very recently, LiDAR-based 4D panoptic segmentation~\cite{ayguen2021cvpr} was formally introduced, and it shares similarities to our targeted task of moving instance segmentation. Thus, we shortly summarize related work targeting this task but emphasize that moving instance tracking is a special case of 4D panoptic segmentation. State-of-the-art approaches~\cite{ayguen2021cvpr,hong2022arxiv,agarwalla2023iros,kreuzberg2022eccvws,zhu2023arxiv} aggregate multiple point clouds and perform instance segmentation within a fused scan. 
Agarwalla~\etalcite{agarwalla2023iros} follow CenterPoint~\cite{yin2021cvpr} and directly predict the velocities of the objects in concatenated point clouds and perform a greedy nearest-neighbor association of the instances. In contrast, Zhu~\etalcite{zhu2023arxiv} extend 4D-Stop~\cite{kreuzberg2022eccvws} and propose to learn the offsets as equivariant vector fields. Despite the tremendous progress, aggregated scan processing comes with a significant computational burden and induces latencies. 

Tracking-by-detection algorithms are the most common approaches~\cite{chen2023cvpr,pang2022eccv,reid1979ac2} to work on a sequential scan basis. These algorithms first obtain object detections in the current frame and associate them across time which can be formulated as bipartite graph matching. The data association is often based on a cost matrix which can be solved by the Hungarian method~\cite{kuhn1955nrlq1} or greedy-matching algorithms~\cite{yin2021cvpr}. The cost matrix is a similarity matrix comparing the existing tracks with the newly identified objects based on appearance or geometric features. To include motion information, 
filtering algorithms~\cite{benbarka2021iros,chiu2020arxiv,weng2020iros} such as the Kalman filter~\cite{kalman1960tasme} utilize real-world physical models to estimate the state transition of instances. Based on the predictions, AB3DMOT~\cite{weng2020iros} calculates the 3D IoU as the cost for the association. However, IoU-based association is inappropriate for radar signal processing because instances comprise single points. Chiu~\etalcite{chiu2020arxiv} enhance performance by utilizing the Mahalanobis distance, and CenterPoint~\cite{yin2021cvpr} performs a center-based greedy matching by predicting object velocities.
Marcuzzi~\etalcite{marcuzzi2022icra} combine appearance and motion cues to associate instances, including class-dependent contrastive learning. CXTrack~\cite{xu2023cvpr} and MotionTrack~\cite{zhang2023cvprws} utilize attention-based similarity features to track single and multi-objects, respectively. However, MotionTrack struggles to associate objects based on attention due to the sparsity of the point clouds, which is more severe for noisy radar data. Additionally, the proposed voxel-based backbones~\cite{marcuzzi2022icra,zhang2023cvprws} induce discretization artifacts, harming accuracy. 

In this paper, we follow recent advancements and propose a novel moving instance tracking method that combines geometric and appearance features for sparse and noisy radar data. Our Radar Tracker includes a temporal offset prediction module to capture important motion information and enhance center-based tracking. Furthermore, our proposed transformer-based network encodes the instance information to derive attention-based association scores incorporating valuable appearance features. Our combined data association overcomes the shortcomings of center-based association and enhances state-of-the-art performance for moving instance tracking in sparse radar point clouds.

\begin{figure*}[t]
 \centering
 \fontsize{8pt}{8pt}\selectfont
 \def\svgwidth{\textwidth}
 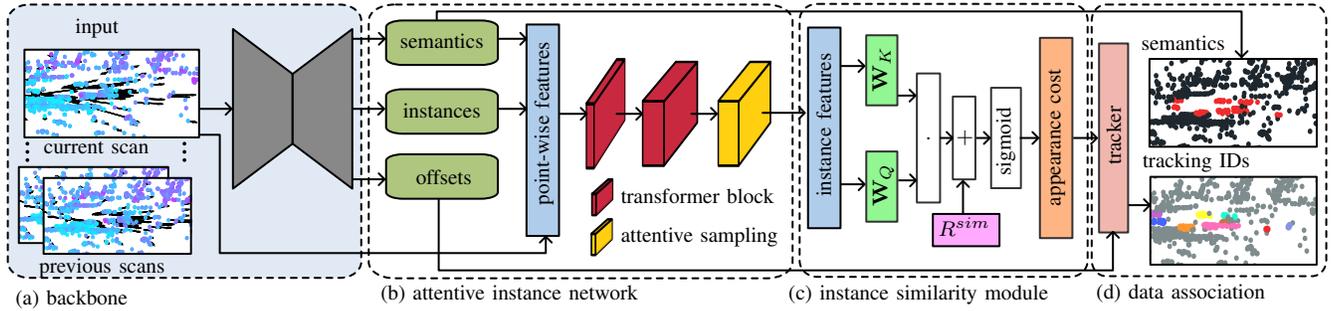%
 \caption{The detailed design of the individual modules of our Radar Tracker. (a) The backbone is extended with the offset predictions and provides the semantic classes and instance predictions. (b) The attentive instance network extracts features to represent instances. (c)~Our instance similarity module determines the appearance-based association matrix to enhance instance tracking. (d) The data association utilizes the appearance and geometric features to predict the tracking IDs~$\mathcal{T}^{\mathrm{track}}$ of moving instances in sparse radar point clouds. }
 \vspace{-0.1cm}
 \label{fig:modules}
  \vspace{-0.2cm}
\end{figure*}
\section{Our Approach to Track Moving Instances}
\label{sec:main}
Our approach aims to achieve reliable moving instance tracking in sparse radar point clouds. We follow the tracking-by-detection paradigm and extend the Radar Instance Transformer~\cite{zeller2023tro} with dedicated tracking modules, as illustrated in~\figref{fig:modules}. We directly predict the temporal offset for each detection within single radar scans to enhance segmentation and enable direct center-based association. We utilize the self-attention mechanism~\cite{vaswani2017nips} to regress an additional cost function and include appearance features to improve tracking. The final association combines geometric and appearance features to enhance scene understanding.

\subsection{Moving Instance Segmentation Backbone}
The performance of the instance segmentation backbone limits the tracking of objects. Therefore, we utilize the state-of-the-art Radar Instance Transformer~\cite{zeller2023tro} as the backbone to extract moving instances reliably. The current radar scan $\mathcal{P}^{t}$ at time~$t$, which comprises the point coordinates and the radar features such as the Doppler velocity and radar cross section, is efficiently enriched with temporal information from $N^p$ previous scans $\mathcal{P}^{t-N^p}$ by the sequential attentive feature encoding module. The processed single scan, including temporally enriched point-wise features, is then passed through the network. The outputs of the backbone are the moving object segmentation~(MOS) labels $\mathcal{S}^{\mathrm{MOS}}$, the instance IDs ${\mathcal{I}=\{I_1,\dots, I_N\}}$ with $I_i\in\NN$, and the point-wise features $\mathbf{X}^b$. We utilize the predictions as input to our Radar Tracker. Since we do not rely on additional information, we can potentially substitute the backbone for other moving instance segmentation networks.

\subsection{Offset Prediction Module}
Geometric features of instances are essential to track moving objects reliably. However, in sparse radar point clouds, the appearance of objects changes, making tracking based on bounding boxes difficult~\cite{weng2020iros}. Especially the case of single-point instances is not covered adequately. Therefore, our Radar Tracker focuses on center-based associations to exploit important geometric properties. 

Furthermore, future state prediction is crucial to associate tracks and objects over time. In contrast to other approaches~\cite{chen2023cvpr,yin2021cvpr}, which predict velocity vectors for bounding boxes or voxels, we directly process the point cloud on a per-point basis to include per-point motion cues. Our resulting approach first utilizes the commonly used offset prediction head~\cite{hong2022arxiv} to regress offsets $\mathbf{O} \in \RR^{N\times 2}$ to the instance center $C^{t}\in \RR^{N \times 2}$ of the current scan. Secondly, we predict the temporal offset $\mathbf{O}^{\mathrm{temp}} \in \RR^{N\times 2}$ for each point, which is a vector that points to the center of the instance $C^{t+1}\in \RR^{N \times 2}$ in the next scan. We calculate the center as the average of the coordinates of the points belonging to the instance.

The input to our offset prediction module is the concatenation of the features of the backbone $\mathbf{X}^b$, and the point coordinates $\mathbf{P}^t$ of the current scan to include fine-grained position information. For the individual offset prediction heads, we combine two fully connected layers, batch normalization~\cite{ioffe2015icml} and a rectified linear unit~(ReLU)~\cite{nair2010icml}. 
The resulting offsets directly incorporate the information for center-based moving instance tracking and add motion cues about moving instances within each scan. 
Additionally, the temporal offset includes a regression target for single-point moving instances, which does not account for the standard offset prediction and enhances segmentation.

\subsection{Attentive Instance Network}
Center-based data association works remarkably well. However, the geometric association often neglects appearance features, which are essential if the geometric features are inaccurate or multiple agents interact, which makes a purely geometrically based approach challenging to solve. In contrast to other methods~\cite{marcuzzi2022icra,chen2023cvpr}, we propose to extract discriminative instance features by a transformer-based network to reduce information loss in sparse radar data.

Our attentive instance network comprises two transformer blocks and an attentive aggregation module. The input consists of the point coordinates $\mathbf{P}^{\text{in}}=[\mathbf{p}_1, \dots, \mathbf{p}_{{N}^{\text{mov}}}]^{\top} \in \RR^{{N}^{\text{mov}}\times 2}$ and the features $\mathbf{X}^{\text{in}}=[\mathbf{x}_1,\dots,\mathbf{x}_{{N}^{\text{mov}}}]^{\top} \in \RR^{{N}^{\text{mov}}\times D}$, where $\mathbf{p}_i\in\RR^{2}$ and $\mathbf{x}_i\in\RR^{D}$ for ${N}^{\text{mov}}$ moving~(mov) points. Hence, $\mathbf{X}^{\text{in}}$ only includes a subset of the features of the backbone $\mathbf{X}^b$, which includes moving and static points. During training, we select the instances based on the ground truth labels and for the inference based on the semantic and instance predictions of the backbone.
We follow the backbone design proposed in Zeller~\etalcite{zeller2023tro} to extract point-wise information. The transformer block is a residual block, including a feature dimension expansion that embeds a transformer layer. We first process the input features~$\mathbf{X}^{\text{in}} \in \RR^{{N}^{\text{mov}}\times D}$ by a linear layer with weight matrix $\mathbf{W}_l \in \RR^{D\times D_1}$ to increase the feature dimension. The resulting features $\mathbf{X}^{\text{in}}_l$ and corresponding point coordinates are fed into a transformer layer. The output features are processed by another linear layer and added to the skip connection features. 
For the transformer layer, we follow the design of the Point Transformer~\cite{zhao2021iccv}. We encode the features of the moving instances $\mathbf{X}^{\text{in}}_l$ as queries $\mathbf{Q}$, keys $\mathbf{K}$, and values $\mathbf{V}$ as follows:
\begin{align}
\mathbf{Q} &= \mathbf{X}^{\text{in}}_l \mathbf{W}_Q\text{,} &
\mathbf{K} &= \mathbf{X}^{\text{in}}_l \mathbf{W}_K\text{,} &
\mathbf{V} &= \mathbf{X}^{\text{in}}_l \mathbf{W}_V\text{,}
\label{eq:1}
\end{align}
where $\mathbf{W}_{Q}$, $\mathbf{W}_{K}$, and $\mathbf{W}_{V} \in \RR^{D_1\times D_1}$ are learned linear projections. To reduce the computational burden, especially for large instances, we restrict the attention mechanism to local areas. We calculate the $k$-nearest neighbors with $k=N^l$ for the points in the current instance. We apply the sample and grouping algorithm~\cite{qi2017nips2} to extract the related queries, keys, and values, resulting in $\mathbf{Q}^{\text{sg}},\mathbf{K}^{\text{sg}}$, and $\mathbf{V}^{\text{sg}} \in \RR^{N^{\text{mov}} \times N^l \times D_1}$. Additionally, we calculate the relative positions $ \mathbf{r}_{i,j}=\mathbf{p}_i-\mathbf{p}_j$ within the local areas of the instances where $\mathbf{p}_i$ and $\mathbf{p}_j \in \mathbf{P}^{\text{in}}$.
We process the relative positions $\mathbf{r}_{i,j}$ by a multi-layer perceptron~(MLP), including two linear layers with weight matrix $\mathbf{W}_1 \in \RR^{2\times 2}$ and $\mathbf{W}_2 \in \RR^{2\times D_1}$, batch normalization~\cite{ioffe2015icml}, and ReLU activation function~\cite{nair2010icml} to derive the relative positional encoding~\cite{zhao2021iccv} $\mathbf{R} \in \RR^{N^{\text{mov}} \times N^l \times D_1}$.

We adopt vector attention~\cite{zhao2020cvpr} and subtract the encoded keys from the encoded queries to calculate the attention weights $\mathbf{A} \in \RR^{N^{\text{mov}} \times N^l \times D_1}$ for the individual points $i$. Additionally, we add the relative positional encoding $\mathbf{R}_{i}$ before we determine the individual weighting of the features by the softmax function as follows:
\begin{align}
\mathbf{A}_{i}= \text{softmax}(\mathbf{Q}^{\text{sg}}_{i}-\mathbf{K}^{\text{sg}}_{i}+\mathbf{R}_{i})\mathrm{.}
\label{eq:4}
\end{align}

To calculate the output features $\mathbf{X}_{1}^{\text{out}} \in \RR^{N^{\text{mov}} \times D_1}$ of the transformer layer, we calculate the sum of the element-wise multiplication, indicated by $\odot$, and add the relative positional encoding as follows:
\begin{align}
\mathbf{X}^{\text{out}}_{1,i} &= \sum_{j=1}^{N^{l}}{ \mathbf{A}_{i,j}\odot (\mathbf{V}^{\text{sg}}_{i,j}+\mathbf{R}_{i,j}) }.
\label{eq:5}
\end{align}

The attentive aggregation module is inspired by the attention mechanism and follows the attentive sampling operation~\cite{zeller2023ral}. We utilize the attentive aggregation module to keep and combine the information of instances within sparse radar point clouds. We process the output features~$\mathbf{X}_{2}^{\text{out}} \in \RR^{N^{\text{mov}} \times D_2}$ of the second transformer blocks by a linear layer with weight matrix $\mathbf{W}^{\text{agg}} \in \RR^{ D_2 \times D_2}$ and softmax activation function. The resulting outputs are our aggregation weights $\mathbf{A}^{\text{agg}}$, which we utilize to weight the $N^{I}$ points within the individual instance. The final instance feature is derived by the summation of the weighted point features, resulting in:
\begin{align}
\mathbf{X}^{\mathrm{inst}}_i &= \sum_{j=1}^{N^{I}}{ \mathbf{A}^{\text{agg}}_{i,j}\odot \mathbf{X}^{\text{out}}_{i,j}}.
\label{eq:15}
\end{align}

The instance-wise feature vectors comprise the information for the data association. Additionally, we extract the coordinates $\mathbf{P}^{\text{inst}}$ of each instance to include position information in the association step.

\subsection{Instance Similarity Module }
The essential part of improving the data association of tracks and newly detected objects is the cost function or similarity measure for the tracking. We utilize the features and center coordinates of our attentive instance network to determine the similarities and incorporate essential appearance features.

We encode the features $\mathbf{X}^{\mathrm{inst}}$ as queries $\mathbf{Q}^{\mathrm{sim}}$ and keys $\mathbf{K}^{\mathrm{sim}}$ following~\eqref{eq:1} and perform dot product attention to derive an attention-based similarity value. Within the attention matrix, we include the self- and cross-attention of the instances. Additionally, we calculate the relative center positions~$\mathbf{r}_{\text{center}}$ where $\mathbf{p}_i$ and $\mathbf{p}_j \in \mathbf{P}_{\text{center}}$ and encode the relative positions as $\mathbf{R}^{\mathrm{sim}}$ by an MLP. We reduce the dimensionality of the encoding to one.

To calculate the resulting attention-based instance similarity scores, we replace the softmax function with an element-wise sigmoid function and add the positional center encoding as follows:
\begin{align}
\mathbf{A}^{\mathrm{sim}}= \text{sigmoid}\left(\mathbf{Q}^{\mathrm{sim}}{\mathbf{K}^{\mathrm{sim}}}^{\top}+\mathbf{R}^{\mathrm{sim}}\right).
\label{eq:dot}
\end{align}

The similarity scores incorporate the feature and position information and directly indicate how likely two instances belong together. To utilize the scores as an additional cost function, we calculate the similarity cost as:
\begin{align}
\mathbf{C}^{\mathrm{sim}} =\frac{1}{(\mathbf{A}^{\mathrm{sim}}+\epsilon)}\text{,}
\label{eq:17}
\end{align}
where $\epsilon$ is an arbitrarily small constant for numerical stability. The similarity cost depends on the appearance features and thus incorporates essential information for moving instance tracking.

\subsection{Data Association}
The central part of our data association is the offset predictions $\mathbf{O}^{\mathrm{temp}}$ and $\mathbf{O}$ for the coordinates of the instance $\mathbf{P}^{\text{inst}}$ because the center-based association within a small distance is reliable for tracking moving instances. However, to improve data association when the geometric association is imprecise, we utilize the attention-based instance similarity scores $\mathbf{A}^{\mathrm{sim}}$ to enhance tracking performance. 

We first calculate the Euclidean distance $d$ based on the center predictions of our method. For the existing tracks, the center is defined by the temporal offset predictions $\mathbf{O}^{\mathrm{temp}}$, whereas for the newly identified instances, the offset $\mathbf{O}$ is utilized. Since a global mapping includes multiple misleading connections, which would be considered in the optimization step, we directly restrict the optimization to local areas. Therefore, we cluster the instances based on the distances $d$ into local areas with DBSCAN~\cite{ester1996kdd2}. After the clustering, we utilize the local cost matrices, \ie only instances in each cluster, and perform Hungarian matching~\cite{kuhn1955nrlq1}. 

Additionally, we process the input features $\mathbf{X}^{\text{in}}$ by our model to extract instance features $\mathbf{X}^{\text{inst}}$ and determine the similarity scores $\mathbf{A}^{\mathrm{sim}}$ of the tracks and the objects. Since the geometric information is precious within the short-range displacement, we determine a threshold $t_{d1}$ where the data association is purely based on the geometric assignment. Above that threshold, we include the similarity cost function $\mathbf{C}^{sim}$ to perform the association. Therefore, we determine a similarity cost threshold $t_{c}$, which resolves whether the object is assigned to the corresponding track. 
The similarity cost threshold also handles occlusion and initializes new tracks. Furthermore, we define a second distance-based threshold $t_{d2}$ where the appearance-based association is difficult, and the association is omitted. We update the existing tracks with the information of the assigned objects. Occluded tracks are propagated according to the temporal offset predictions $\mathbf{O}^{\mathrm{temp}}$, and we initialize new tracks with the corresponding information of the object. The data association incorporates geometric and appearance features to enhance tracking performance.

\subsection{Implementation Details}
We implemented our approach in PyTorch~\cite{paszke2019nips} and trained the instance segmentation backbone and our Radar Tracker with one Nvidia A100 GPU. We adopt the training parameter of the Radar Instance Transformer~\cite{zeller2023tro}. To learn the standard and temporal offsets of the radar detections, we add for both offsets the following loss function:
\begin{equation}
    L^{\mathrm{offset}} = \frac{1}{N}\sum_{i=1}^{N}\lVert \mathbf{o}_i - (\mathbf{c}_i - \mathbf{p}_i) \rVert_1\text{,}
\end{equation}
where $N$ is the number of points in the point cloud, and $\mathbf{c}_i$ is the respective center of the instance that $\mathbf{p}_i$ belongs to.

We utilize the AdamW~\cite{loshchilov2017iclr} optimizer with an initial learning rate of 0.001 to train our Radar Tracker. We process the original features with dimension $D=4$, comprising the point coordinates $(x^C_i, y^C_i)$, the radar cross section $\sigma_i$, and the ego-motion compensated Doppler velocity $v_i$, by the transformer blocks where $D_1=64$ and $D_2=256$. We define the local areas for sample and grouping by $N_l=6$ points. The attentive aggregation module keeps the feature dimension and combines the information within one feature vector. 
The batch of our method includes 64 scan pairs where only the first scan is considered during the loss calculation. Hence, the network is able to predict the data association and if the objects within the first scan are present in the second scan. We supervise the attentive similarity output by a binary cross entropy loss. 

We set the bandwidth $b=10$ for the clustering using DBSCAN to determine local areas for the association. We set the distance based thresholds $t_{d1}=\SI{5}{m}$ and $t_{d2}=\SI{10}{m}$. We keep the tracks for 12 consecutive scans. The cost threshold for the attentive similarity is set to $t_{c}=1.5$. 
We add points belonging to the static class as additional instances for data augmentation to include the differentiation between static and moving points in the attentive similarity. 

\begin{table}[t]
    \centering
\begin{tabular}{@{}lccc@{}}
\toprule
Approach           & $\mathrm{LSTQ}$ & $\text{S}_{\text{assoc}}$ & $\text{S}_{\text{cls}}$ \\ \midrule
MOT~\cite{weng2020iros}             & 42.4 & 19.4 &  \textbf{92.7} \\
Center tracking~\cite{yin2021cvpr} + Hungarian~\cite{kuhn1955nrlq1} & 59.3 & 38.0 &  \textbf{92.7} \\
CA-Net~\cite{marcuzzi2022icra}          & 34.8 & 13.0 &  \textbf{92.7} \\ 
Ours            & \textbf{66.8} & \textbf{48.2} & \textbf{92.7} \\ \bottomrule
\end{tabular}%

\caption{Moving instance tracking results on the RadarScenes test set in terms of $\mathrm{LSTQ}$, $\text{S}_{\text{cls}}$, and $\text{S}_{\text{assoc}}$ scores. 
  \vspace{-0.4cm}}
  \label{tab:resall}
\end{table}
\section{Experimental Evaluation}
\label{sec:exp}
The main focus of this work is to enable reliable moving instance tracking in sparse and noisy radar point clouds.
We present our experiments to show the capabilities of our method and to support our key claims, which include that our method outperforms existing state-of-the-art methods in moving instance tracking. Secondly, our temporal offset prediction enhances the classification and tracking score by adding additional motion cues. Thirdly, our attention-based association scores incorporate valuable appearance features enhancing performance.

\subsection{Experimental Setup}
\label{sec:expset}
We train and evaluate our model on the RadarScenes~\cite{schumann2021icif} dataset since it is the only large-scale high-resolution radar dataset~\cite{zhou2022sensors} that includes per-point annotations for moving instance tracking under versatile scenarios. We follow Zeller~\etalcite{zeller2023icra} and split the 158 sequences into 130 sequences for training, 6 for validation, and 22 for testing. We perform the ablation studies on the validation set. We merge the data of the four individual radar sensors to obtain information about the surroundings of the vehicle~\cite{zeller2023icra}.

We utilize the LiDAR segmentation and tracking quality~($\mathrm{LSTQ}$) score~\cite{ayguen2021cvpr} to evaluate the moving instance tracking performance. The LSTQ is designed for evaluating point-based segmentation and tracking methods and does not depend on LiDAR-specific properties. Additionally, the adaptation of the metric enables comparability for follow-up research. The LSTQ combines the classification score~$\text{S}_{\text{cls}}$ and association score~$\text{S}_{\text{assoc}}$, resulting in $\text{LSTQ} = \sqrt{ \text{S}_{\text{cls}} \times \text{S}_{\text{assoc}}}$.

\subsection{Moving Instance Tracking}
\label{sec:mit}
The first experiment evaluates the performance of our approach and its outcomes support the claim that our method achieves state-of-the-art performance in moving instance tracking in sparse and noisy radar scans. We compare our Radar Tracker to the high-performing networks with strong performance in point-based tracking benchmarks. However, we do not consider the best performing Eq-4D-StOP~\cite{zhu2023arxiv} since it incorporates large rotations of the input point clouds, which is detrimental to radar data~\cite{palffy2022ral}. Therefore, we utilize CA-Net~\cite{marcuzzi2022icra}, MOT~\cite{weng2020iros}, and the center tracking approach proposed by Yin~\etalcite{yin2021cvpr} as baselines. We extend the center tracking with Hungarian matching~\cite{kuhn1955nrlq1} and directly use the measured Doppler velocities to perform the tracking instead of predicting the velocities of the individual bounding boxes. For the MOT~\cite{weng2020iros} approach, we utilize the IoU as the cost to illustrate the limitations. We adopt the Radar Instance Transformer~\cite{zeller2023tro} as the backbone for all methods since it is the best-performing approach for moving instance segmentation and thus enables a fair comparison.

Our Radar Tracker outperforms the existing methods, especially in terms of $\mathrm{LSTQ}$ and $\text{S}_{\text{assoc}}$, as displayed in~\tabref{tab:resall}. As mentioned, the MOT~\cite{weng2020iros} approach struggles to associate small instances due to the IoU-based association. The center-based tracking overcomes these limitations and enhances performance. Nevertheless, both methods neglect the appearance features of the instances and thus can not compensate for the shortcomings of geometric tracking. CA-Net combines both features within one cost function. However, the method struggles to capture instance information and to associate the instances based on appearance features. We argue that extracting appropriate features is challenging in sparse radar data, and the design of the network and association function is crucial to enhance accuracy. Our method exceeds these limits and reliably tracks moving instances by combining geometric and appearance features.

\begin{table}[t]
  \centering
\begin{tabular}{@{}lccc@{}}
\toprule
Approach      & standard offset & temporal offset & $\mathrm{IoU}^{\mathrm{mov}}$  \\ \midrule
RIT~\cite{zeller2023tro}        & &                & 84.4 \\
Ours      & {\checkmark }        &                 & 85.2 \\
Ours      &        &{\checkmark }                  & 85.3 \\
Ours  & {\checkmark }        & {\checkmark }                 & \textbf{85.4} \\ \bottomrule
\end{tabular}%

\caption{Influence of the temporal and standard offset predictions in terms of $\mathrm{IoU}^{\mathrm{mov}}$ on the RadarScenes validation set.}
  \label{tab:offsetreg}
  \vspace{-0.4cm}
\end{table}

\subsection{Ablation Study on Offset Predictions}
\label{sec:abloff}
The second experiment evaluates our offset predictions, especially the temporal offset, and illustrates that our approach is capable of including valuable motion cues to enhance segmentation and tracking quality. To evaluate the segmentation performance, we utilize the $\mathrm{IoU}^{\mathrm{mov}}$ since segmentation of moving detection is essential for tracking. We extend the backbone, the Radar Instance Transformer, with the standard offset prediction and the temporal offset prediction as additional regression targets, as depicted in~\tabref{tab:offsetreg}. The standard offset, which points to the center of the instance within the current scan, already improves the $\mathrm{IoU}^{\mathrm{mov}}$ by 0.8 absolute percentage points. Despite that improvement, the temporal offset prediction enhances the performance by an additional 0.1 absolute percentage point. We presume that the temporal offset prediction includes stronger motion cues, especially for instances comprising single detection that do not have a regression target for the standard offset. We combine both offset predictions to enable direct center-based tracking and achieve the best $\mathrm{IoU}^{\mathrm{mov}}$ of 85.4\%. 

To verify that the offset predictions improve the tracking performance, we evaluate a simple center-based association with and without the center predictions. We remove the appearance features to strictly assess the performance of the geometric approach. The offset prediction improves the $\text{S}_{\text{assoc}}$ from 49.5\% to 50.2\%, which underlines the advantage of direct temporal offset predictions.

\subsection{Ablation Study on Attentive Association}
\label{sec:ablmodel}
Finally, we analyze our method concerning the ability to extract reliable attentive similarity scores to associate instances. Therefore, we evaluate the different components of our method as detailed in~\tabref{tab:ablatten}. First, we remove the positional encoding within our attentive instance association, which results in a decrease of 0.3 absolute percentage points. We argue that positional encoding is important to differentiate between similar instances within the scan. 

In the second step, we add an additional no-object regression target~\cite{zhang2023cvprws} to the attention score to address the occlusion within the appearance features. However, distant instances are often detected in one scan but not covered in the next one, leading to several no-object assignments as ground truth. We assume that this forces the network to assign more instances to the no object class, and the information to track the instances is not covered adequately, which results in a 0.2 absolute percentage points decrease of $\text{S}_{\text{assoc}}$. Additionally, we tried to remove the sigmoid function~\cite{zhang2023cvprws} to directly learn the attention scores. However, this also results in a decrease in performance. 
To verify that the association based on the attention scores enhances accuracy, we evaluate our method, including only geometric information for the threshold $t_{d2}=\SI{10}{m}$. The geometric association performs worse compared to our combined approach.
Hence, the appearance features are essential to track the instances and resolve ambiguities within larger distances.
In summary, our evaluation suggests that our method
provides competitive moving instance tracking results in sparse and noisy radar point clouds by incorporating geometric and appearance features. Thus, we supported
all our claims with this experimental evaluation.
\begin{table}[t]
\centering%
\begin{tabular}{@{}lc@{}}
\toprule
Approach                        & $\text{S}_{\text{assoc}}$ \\ \midrule
Ours w/o positional encoding & 54.0   \\
Ours with no object class    & 54.1   \\
Ours w/o sigmoid             & 54.0   \\ 
Geometric association $t_{d2}=\SI{10}{m}$ & 52.2   \\ 
Ours                   & \textbf{54.3} \\\bottomrule
\end{tabular}%
\caption{Influence of the design decision for the attentive association on the RadarScenes validation set.}
  \label{tab:ablatten}
  \vspace{-0.6cm}
\end{table}

\section{Conclusion}
\label{sec:conclusion}

In this paper, we presented a novel approach for moving instance tracking in sparse and noisy radar point clouds. Our method exploits temporal offset predictions to encode geometric information to enhance segmentation and tracking. We incorporate appearance features and introduce an attention-based association cost to improve the tracking quality. 
This allows us to successfully associate individual instances based on valuable geometric and appearance features over time. We furthermore evaluated our method on the radar moving instance tracking benchmark based on the RadarScenes dataset, providing comparisons to other methods and supporting all claims made in this paper. 
The experiments suggest that combining geometric and appearance features is essential to achieve good performance on moving instance tracking in sparse radar data. 
Overall, our approach outperforms the state-of-the-art methods, taking a step towards reliable moving instance tracking and sensor redundancy for autonomous vehicles.

\bibliographystyle{plain_abbrv}

\bibliography{glorified,new}

\end{document}

%% file: stachnisslab-latex.tex
\usepackage{graphics}           
\usepackage{times}              
\usepackage{amsmath}            
\usepackage{amssymb}            
\usepackage{graphicx}
\usepackage{algorithm}
\usepackage[noend]{algpseudocode}
\usepackage{booktabs}
\usepackage{color}
\usepackage[nocompress]{cite} %
\definecolor{instructioncolor}{rgb}{.5,.5,.5}

\usepackage[font=small]{caption}

\def\figref#1{Fig.~\ref{#1}}
\def\tabref#1{Tab.~\ref{#1}}
\def\eqref#1{Eq.~(\ref{#1})}

\makeatletter
\usepackage{xspace}
\DeclareRobustCommand\onedot{\futurelet\@let@token\@onedot}
\def\@onedot{\ifx\@let@token.\else.\null\fi\xspace}
 
\def\ie{i.e\onedot}

\def\etal{{et al}\onedot}
\makeatother

\def\etalcite#1{\etal~\cite{#1}}

\usepackage{array}
\newcolumntype{L}[1]{>{\raggedright\let\newline\\\arraybackslash\hspace{0pt}}m{#1}}
\newcolumntype{C}[1]{>{\centering\let\newline\\\arraybackslash\hspace{0pt}}m{#1}}
\newcolumntype{R}[1]{>{\raggedleft\let\newline\\\arraybackslash\hspace{0pt}}m{#1}}

%% file: stachnisslab-math.tex
\newcommand{\RR}{\mathbb{R}}
\newcommand{\NN}{\mathbb{N}}

%% file: pics/motivation_radartrack.pdf_tex
\begingroup%
  \makeatletter%
  \providecommand\color[2][]{%
    \errmessage{(Inkscape) Color is used for the text in Inkscape, but the package 'color.sty' is not loaded}%
    \renewcommand\color[2][]{}%
  }%
  \providecommand\transparent[1]{%
    \errmessage{(Inkscape) Transparency is used (non-zero) for the text in Inkscape, but the package 'transparent.sty' is not loaded}%
    \renewcommand\transparent[1]{}%
  }%
  \providecommand\rotatebox[2]{#2}%
  \newcommand*\fsize{\dimexpr\f@size pt\relax}%
  \newcommand*\lineheight[1]{\fontsize{\fsize}{#1\fsize}\selectfont}%
  \ifx\svgwidth\undefined%
    \setlength{\unitlength}{245.71800232bp}%
    \ifx\svgscale\undefined%
      \relax%
    \else%
      \setlength{\unitlength}{\unitlength * \real{\svgscale}}%
    \fi%
  \else%
    \setlength{\unitlength}{\svgwidth}%
  \fi%
  \global\let\svgwidth\undefined%
  \global\let\svgscale\undefined%
  \makeatother%
  \begin{picture}(1,0.84649882)%
    \lineheight{1}%
    \setlength\tabcolsep{0pt}%
    \put(0,0){\includegraphics[width=\unitlength,page=1]{pics/motivation_radartrack.pdf}}%
    \put(0.13258814,0.07373512){\color[rgb]{0,0,0}\rotatebox{90}{\makebox(0,0)[lt]{\lineheight{1.25}\smash{\begin{tabular}[t]{l}time t+1\end{tabular}}}}}%
    \put(0.07054361,0.06954855){\color[rgb]{0,0,0}\rotatebox{90}{\makebox(0,0)[lt]{\lineheight{1.25}\smash{\begin{tabular}[t]{l}instances\end{tabular}}}}}%
    \put(0,0){\includegraphics[width=\unitlength,page=2]{pics/motivation_radartrack.pdf}}%
    \put(0.33504971,0.26897966){\color[rgb]{0,0,0}\makebox(0,0)[lt]{\lineheight{1.25}\smash{\begin{tabular}[t]{l}consistent tracking over time\end{tabular}}}}%
    \put(0.13285872,0.37453902){\color[rgb]{0,0,0}\rotatebox{90}{\makebox(0,0)[lt]{\lineheight{1.25}\smash{\begin{tabular}[t]{l} time t\end{tabular}}}}}%
    \put(0.07216469,0.34933462){\color[rgb]{0,0,0}\rotatebox{90}{\makebox(0,0)[lt]{\lineheight{1.25}\smash{\begin{tabular}[t]{l}instances\end{tabular}}}}}%
    \put(0.12820407,0.6479862){\color[rgb]{0,0,0}\rotatebox{90}{\makebox(0,0)[lt]{\lineheight{1.25}\smash{\begin{tabular}[t]{l} time t\end{tabular}}}}}%
    \put(0.06757257,0.62337913){\color[rgb]{0,0,0}\rotatebox{90}{\makebox(0,0)[lt]{\lineheight{1.25}\smash{\begin{tabular}[t]{l}semantics\end{tabular}}}}}%
    \put(0,0){\includegraphics[width=\unitlength,page=3]{pics/motivation_radartrack.pdf}}%
  \end{picture}%
\endgroup%

%% file: pics/radartracker_architecture.pdf_tex
\begingroup%
  \makeatletter%
  \providecommand\color[2][]{%
    \errmessage{(Inkscape) Color is used for the text in Inkscape, but the package 'color.sty' is not loaded}%
    \renewcommand\color[2][]{}%
  }%
  \providecommand\transparent[1]{%
    \errmessage{(Inkscape) Transparency is used (non-zero) for the text in Inkscape, but the package 'transparent.sty' is not loaded}%
    \renewcommand\transparent[1]{}%
  }%
  \providecommand\rotatebox[2]{#2}%
  \newcommand*\fsize{\dimexpr\f@size pt\relax}%
  \newcommand*\lineheight[1]{\fontsize{\fsize}{#1\fsize}\selectfont}%
  \ifx\svgwidth\undefined%
    \setlength{\unitlength}{505.89001465bp}%
    \ifx\svgscale\undefined%
      \relax%
    \else%
      \setlength{\unitlength}{\unitlength * \real{\svgscale}}%
    \fi%
  \else%
    \setlength{\unitlength}{\svgwidth}%
  \fi%
  \global\let\svgwidth\undefined%
  \global\let\svgscale\undefined%
  \makeatother%
  \begin{picture}(1,0.23325228)%
    \lineheight{1}%
    \setlength\tabcolsep{0pt}%
    \put(0,0){\includegraphics[width=\unitlength,page=1]{pics/radartracker_architecture.pdf}}%
    \put(0.03366036,0.02821546){\color[rgb]{0,0,0}\makebox(0,0)[lt]{\lineheight{1.25}\smash{\begin{tabular}[t]{l}previous scans\end{tabular}}}}%
    \put(0.03715217,0.11780203){\color[rgb]{0,0,0}\makebox(0,0)[lt]{\lineheight{1.25}\smash{\begin{tabular}[t]{l}current scan\end{tabular}}}}%
    \put(0.85785572,0.19460973){\color[rgb]{0,0,0}\makebox(0,0)[lt]{\lineheight{1.25}\smash{\begin{tabular}[t]{l}semantics\end{tabular}}}}%
    \put(0.85863642,0.10683861){\color[rgb]{0,0,0}\makebox(0,0)[lt]{\lineheight{1.25}\smash{\begin{tabular}[t]{l}tracking IDs\end{tabular}}}}%
    \put(0.06060779,0.20790158){\color[rgb]{0,0,0}\makebox(0,0)[lt]{\lineheight{1.25}\smash{\begin{tabular}[t]{l}input\end{tabular}}}}%
    \put(0,0){\includegraphics[width=\unitlength,page=2]{pics/radartracker_architecture.pdf}}%
    \put(0.3055963,0.1441104){\color[rgb]{0,0,0}\makebox(0,0)[lt]{\lineheight{1.25}\smash{\begin{tabular}[t]{l}instances\end{tabular}}}}%
    \put(0,0){\includegraphics[width=\unitlength,page=3]{pics/radartracker_architecture.pdf}}%
    \put(0.31524069,0.09483988){\color[rgb]{0,0,0}\makebox(0,0)[lt]{\lineheight{1.25}\smash{\begin{tabular}[t]{l}offsets\end{tabular}}}}%
    \put(0,0){\includegraphics[width=\unitlength,page=4]{pics/radartracker_architecture.pdf}}%
    \put(0.62569343,0.08527){\color[rgb]{0,0,0}\rotatebox{90}{\makebox(0,0)[lt]{\lineheight{1.25}\smash{\begin{tabular}[t]{l}instance features\end{tabular}}}}}%
    \put(0.46851094,0.08007096){\color[rgb]{0,0,0}\makebox(0,0)[lt]{\lineheight{1.25}\smash{\begin{tabular}[t]{l}transformer block\end{tabular}}}}%
    \put(0.46860794,0.05104227){\color[rgb]{0,0,0}\makebox(0,0)[lt]{\lineheight{1.25}\smash{\begin{tabular}[t]{l}attentive sampling\end{tabular}}}}%
    \put(0,0){\includegraphics[width=\unitlength,page=5]{pics/radartracker_architecture.pdf}}%
    \put(0.66755613,0.16375865){\color[rgb]{0,0,0}\rotatebox{90}{\makebox(0,0)[lt]{\lineheight{1.25}\smash{\begin{tabular}[t]{l}$\mathbf{W}_{K}$\end{tabular}}}}}%
    \put(0,0){\includegraphics[width=\unitlength,page=6]{pics/radartracker_architecture.pdf}}%
    \put(0.66774662,0.07787322){\color[rgb]{0,0,0}\rotatebox{90}{\makebox(0,0)[lt]{\lineheight{1.25}\smash{\begin{tabular}[t]{l}$\mathbf{W}_{Q}$\end{tabular}}}}}%
    \put(0,0){\includegraphics[width=\unitlength,page=7]{pics/radartracker_architecture.pdf}}%
    \put(0.72974501,0.12314433){\color[rgb]{0,0,0}\rotatebox{90}{\makebox(0,0)[lt]{\lineheight{1.25}\smash{\begin{tabular}[t]{l}$+$\end{tabular}}}}}%
    \put(0,0){\includegraphics[width=\unitlength,page=8]{pics/radartracker_architecture.pdf}}%
    \put(0.76001956,0.10125517){\color[rgb]{0,0,0}\rotatebox{90}{\makebox(0,0)[lt]{\lineheight{1.25}\smash{\begin{tabular}[t]{l}sigmoid\end{tabular}}}}}%
    \put(0,0){\includegraphics[width=\unitlength,page=9]{pics/radartracker_architecture.pdf}}%
    \put(0.84226453,0.10932479){\color[rgb]{0,0,0}\rotatebox{90}{\makebox(0,0)[lt]{\lineheight{1.25}\smash{\begin{tabular}[t]{l}tracker\end{tabular}}}}}%
    \put(0.82412517,0.00911021){\color[rgb]{0,0,0}\makebox(0,0)[lt]{\lineheight{1.25}\smash{\begin{tabular}[t]{l}(d) data association\end{tabular}}}}%
    \put(0,0){\includegraphics[width=\unitlength,page=10]{pics/radartracker_architecture.pdf}}%
    \put(0.30339911,0.19633572){\color[rgb]{0,0,0}\makebox(0,0)[lt]{\lineheight{1.25}\smash{\begin{tabular}[t]{l}semantics\end{tabular}}}}%
    \put(0,0){\includegraphics[width=\unitlength,page=11]{pics/radartracker_architecture.pdf}}%
    \put(0.01601473,0.00432176){\color[rgb]{0,0,0}\makebox(0,0)[lt]{\lineheight{1.25}\smash{\begin{tabular}[t]{l}(a)  backbone\end{tabular}}}}%
    \put(0.28899993,0.00795114){\color[rgb]{0,0,0}\makebox(0,0)[lt]{\lineheight{1.25}\smash{\begin{tabular}[t]{l}(b)  attentive instance network\end{tabular}}}}%
    \put(0.59398127,0.00835412){\color[rgb]{0,0,0}\makebox(0,0)[lt]{\lineheight{1.25}\smash{\begin{tabular}[t]{l}(c)  instance similarity module\end{tabular}}}}%
    \put(0.70590417,0.05596626){\color[rgb]{0,0,0}\makebox(0,0)[lt]{\lineheight{1.25}\smash{\begin{tabular}[t]{l} $R^{sim}$\end{tabular}}}}%
    \put(0,0){\includegraphics[width=\unitlength,page=12]{pics/radartracker_architecture.pdf}}%
    \put(0.79715474,0.07851261){\color[rgb]{0,0,0}\rotatebox{90}{\makebox(0,0)[lt]{\lineheight{1.25}\smash{\begin{tabular}[t]{l}appearance cost\end{tabular}}}}}%
    \put(0,0){\includegraphics[width=\unitlength,page=13]{pics/radartracker_architecture.pdf}}%
    \put(0.41478025,0.07913207){\color[rgb]{0,0,0}\rotatebox{90}{\makebox(0,0)[lt]{\lineheight{1.25}\smash{\begin{tabular}[t]{l}point-wise features\end{tabular}}}}}%
    \put(0.70245087,0.12661773){\color[rgb]{0,0,0}\rotatebox{90}{\makebox(0,0)[lt]{\lineheight{1.25}\smash{\begin{tabular}[t]{l}$\cdot$\end{tabular}}}}}%
    \put(0,0){\includegraphics[width=\unitlength,page=14]{pics/radartracker_architecture.pdf}}%
  \end{picture}%
\endgroup%